# Brain Tumor Sequence Registration with Non-iterative Coarse-to-fine Networks and Dual Deep Supervision

Mingyuan Meng[1], Lei Bi[1], Dagan Feng[1,2], and Jinman Kim[1]

[1] School of Computer Science, The University of Sydney, Sydney, Australia.
[2] Med-X Research Institute, Shanghai Jiao Tong University, Shanghai, China.
`lei.bi@sydney.edu.au`

**Abstract.** In this study, we focus on brain tumor sequence registration between pre-operative and follow-up Magnetic Resonance Imaging (MRI) scans of brain glioma patients, in the context of Brain Tumor Sequence Registration challenge (BraTS-Reg 2022). Brain tumor registration is a fundamental requirement in brain image analysis for quantifying tumor changes. This is a challenging task due to large deformations and missing correspondences between pre-operative and follow-up scans. For this task, we adopt our recently proposed Non-Iterative Coarse-to-finE registration Networks (NICE-Net) – a deep learning-based method for coarse-to-fine registering images with large deformations. To overcome missing correspondences, we extend the NICE-Net by introducing dual deep supervision, where a deep self-supervised loss based on image similarity and a deep weakly-supervised loss based on manually annotated landmarks are deeply embedded into the NICE-Net. At the BraTS-Reg 2022, our method achieved a competitive result on the validation set (mean absolute error: 3.387) and placed 4[th] in the final testing phase (Score: 0.3544).

## 1 Introduction

Deformable image registration aims to establish a dense, non-linear correspondence between a pair of images, which is a crucial step in a variety of clinical tasks such as organ atlas creation and tumor monitoring [1]. Due to pathological changes (e.g., tumor growth) or inter-patient anatomy variations, medical images usually carry many non-linear local deformations. Therefore, unlike the commonly used natural image registration tasks such as photo stitching [2], medical image analysis heavily relies on deformable image registration. However, deformable image registration is still an intractable problem due to image shape and appearance variations [3], especially for image pairs containing pathology-affected tissue changes.

To establish a fair benchmark environment for deformable registration methods, Baheti et al. [4] organized a Brain Tumor Sequence Registration challenge (BraTS-Reg 2022), focusing on brain tumor sequence registration between pre-operative and follow-up Magnetic Resonance Imaging (MRI) scans of brain glioma patients. Brain tumor registration is clinically important as it can advance the understanding of glio-mas and aids in analyzing the tissue resulting in tumor relapse [5]. However, this is a challenging task because: (i) brain tumors usually cause large deformations in brain anatomy, and (ii) there are missing correspondences between the tumor in the pre-operative scan and the resection cavity in the follow-up scan [4].

Traditional methods attempt to solve deformable registration as an iterative optimization problem [6], which is usually time-consuming and has inspired a tendency toward faster deep registration methods based on deep learning [7]. To register image pairs with large deformations, coarse-to-fine deep registration methods were widely used and are regarded as the state-of-the-art [8-13]. Typically, coarse-to-fine registration was implemented by iteratively warping an image with multiple cascaded networks [8, 9] or with multiple network iterations [10]. Recently, non-iterative coarse-to-fine registration methods were proposed to perform coarse-to-fine registration with a single network in a single iteration [11-13], which have demonstrated state-of-the-art registration accuracy even when compared with methods using multiple cascaded networks or multiple network iterations.

In this study, we adopt our recently proposed deep registration method – Non-Iterative Coarse-to-finE registration Networks (NICE-Net) [13]. The NICE-Net performs multiple steps of coarse-to-fine registration in a single network iteration, which is optimized for image pairs with large deformations and has shown state-of-the-art performance on inter-patient brain MRI registration [13]. However, the NICE-Net was not optimized for brain tumor registration with missing correspondences. We extend the NICE-Net by introducing dual deep supervision, where a deep self-supervised loss based on image similarity and a deep weakly-supervised loss based on manually annotated landmarks (ground truth) are deeply embedded into each coarse-to-fine registration step of the NICE-Net. The deep self-supervised loss can leverage the information within image appearance (e.g., texture and intensity pattern), while the deep weakly-supervised loss can leverage ground truth information to overcome the missing correspondences between pre-operative and follow-up scans.

## 2 Materials and Methods

### 2.1 Dataset and Data Preprocessing

The BraTS-Reg 2022 dataset was curated from multiple institutions, containing pairs of pre-operative and follow-up brain MRI scans of the same patient diagnosed and treated for gliomas. The training set contains 140 pairs of multi-parametric MRI scans along with landmark annotations. The multi-parametric MRI sequences include native T1-weighted (T1), contrast-enhanced T1-weighted (T1CE), T2-weighted (T2), and T2 Fluid Attenuated Inversion Recovery (FLAIR) images. The landmark annotations include the landmark coordinates in the baseline scan and their corresponding coordinates in the follow-up scan, which were manually annotated by clinical experts and are regarded as ground truth (refer to [4] for more details about landmark annotations). The validation set contains 20 pairs of MRI scans and was provided to validate the developed registration methods. Finally, the registration methods are submitted in a containerized form to be evaluated on a hidden testing set.

We preprocessed the provided images with the following steps: (i) we cropped and padded each MRI image from 240×240×155 to 144×192×160 with the coordinates of (48:192, 32:224, -5:155), and (ii) we normalized each MRI image within the range of 0 and 1 through min-max normalization.

## 2.2  Non-iterative Coarse-to-fine Registration Networks

Image registration aims to find a spatial transformation $\phi$ that warps a moving image $I_m$ to a fixed image $I_f$, so that the warped image $I_m \circ \phi$ is spatially aligned with the fixed image $I_f$. In this study, the $I_m$ and $I_f$ are two volumes defined in a 3D spatial domain $\Omega \subset \mathbb{R}^3$, and the $\phi$ is parameterized as a displacement field [14]. Our method is based on our recently proposed NICE-Net (Non-Iterative Coarse-to-finE registration Networks) [13]. The architecture of the NICE-Net is shown in Fig. 1, which consists of a feature learning encoder and a coarse-to-fine registration decoder. We employed the NICE-Net to perform four steps of coarse-to-fine registration. At the $i^{th}$ step for $i \in \{1, 2, 3, 4\}$, a transformation $\phi_i$ is produced, with the $\phi_1$ as the coarsest transformation and the $\phi_4$ as the finest transformation. We created two image pyramids as the input, downsampling the $I_f$ and $I_m$ with trilinear interpolation by a factor of $0.5^{(4-i)}$ to obtain $I_f^i$ and $I_m^i$ for $i \in \{1, 2, 3, 4\}$ with $I_f^4 = I_f$ and $I_m^4 = I_m$. In addition, the coordinates of the landmark in the $I_f$ and $I_m$ are denoted as $L_f$ and $L_m$. Along with the downsampling of $I_f$ and $I_m$, the $L_f$ and $L_m$ were also downsampled as $L_f^i$ and $L_m^i$ for $i \in \{1, 2, 3, 4\}$ with $L_f^4 = L_f$ and $L_m^4 = L_m$.

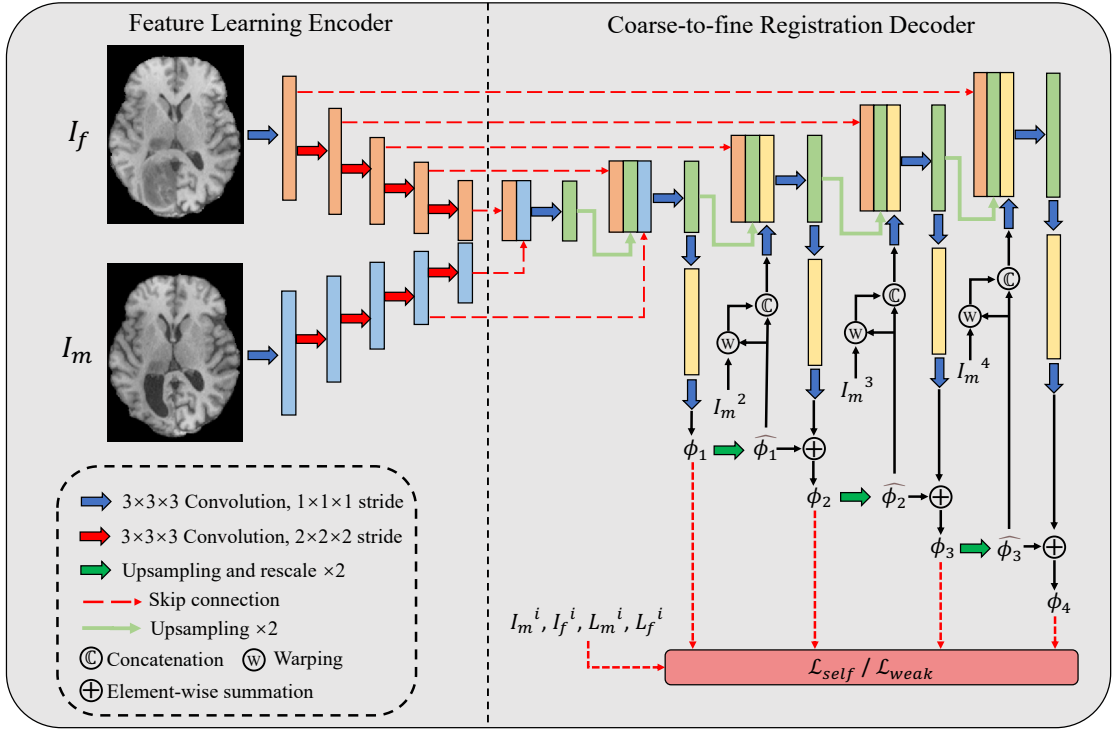

**Fig. 1.** Architecture of the NICE-Net that performs four steps of coarse-to-fine registration.

The feature learning encoder has two paths to separately extract features from the $I_f$ and $I_m$, which is different from the deep registration methods that learn coupled features from the concatenated $I_f$ and $I_m$ [8, 9, 14, 15]. Specifically, the encoder has two identical, weight-shared paths that take $I_f$ and $I_m$ as input. Each path consists of five successive $3 \times 3 \times 3$ convolutional layers, followed by LeakyReLU activation with parameter 0.2. Except for the first convolutional layer, each convolutional layer has a stride of 2 to reduce the resolution of feature maps.

The coarse-to-fine registration decoder performs four steps of registration in a coarse-to-fine manner. Specifically, the decoder has five successive $3 \times 3 \times 3$ convolutional layers to cumulate features from different sources, followed by LeakyReLU activation with parameter 0.2. Except for the last convolutional layer, an upsampling layer is used after each convolutional layer to increase the resolution of feature maps by a factor of 2. The first convolutional layer is used to cumulate features from the encoder, while the first (coarsest) registration step is performed at the second convolutional layer and produces the $\phi_1$. The $\phi_1$ is upsampled by a factor of 2 (as $\widehat{\phi_1}$) and then warps the $I_m{}^2$. The warped image $I_m{}^2 \circ \widehat{\phi_1}$ and the $\widehat{\phi_1}$ are fed into a convolutional layer to extract features and then are leveraged at the second registration step. The second registration step produces a displacement field based on the cumulated features at the third convolution layer, and voxel-wisely add the displacement field to $\widehat{\phi_1}$ to obtain $\phi_2$. We repeat this process until $\phi_4$ is obtained.

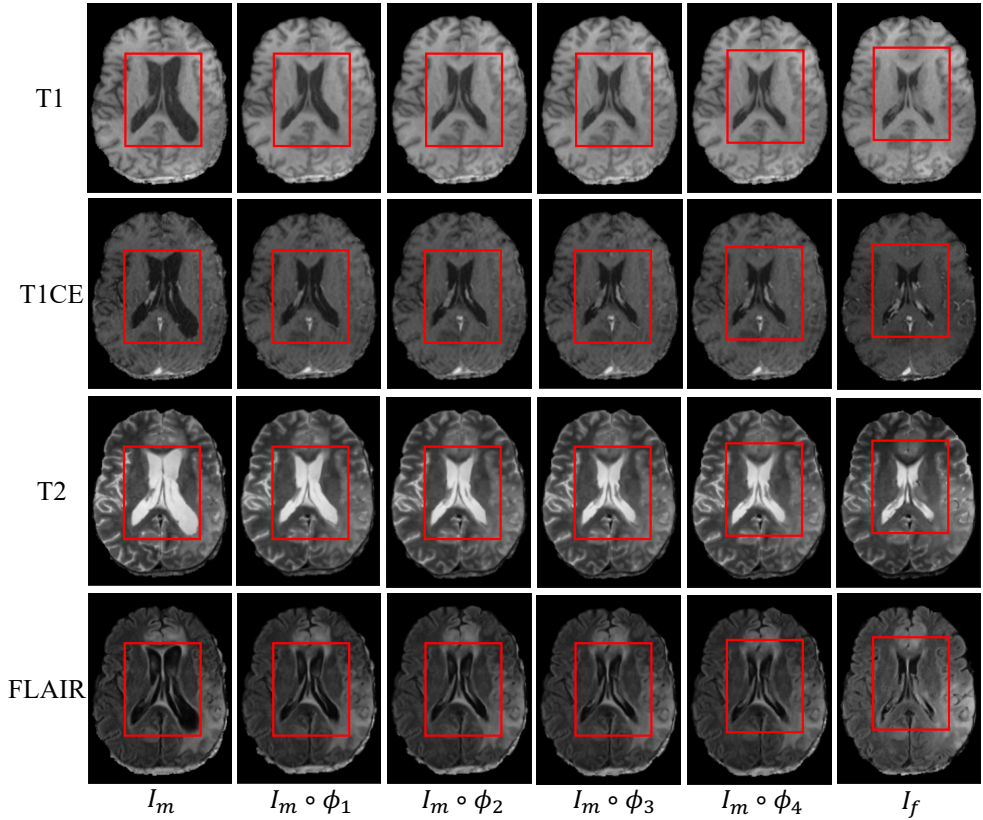

**Fig. 2.** Exemplified registration results with NICE-Net. Red boxes highlight the regions with major changes, gradually being closer to the fixed image $I_f$ after each registration step. The same red bounding box has been placed at the same location for better visual comparison.

During training, all the output $\phi_1$, $\phi_2$, $\phi_3$, and $\phi_4$ are supervised by a deep self-supervised loss $\mathcal{L}_{self}$ and/or a deep weakly-supervised loss $\mathcal{L}_{weak}$ (detailed in Section 2.3 and Section 2.4). During inference, only $\phi_4$ is used to warp the $I_m$ to align with the $I_f$. We present exemplified registration results of the NICE-Net in Fig. 2, which shows that the NICE-Net can perform coarse-to-fine registration to make the moving image $I_m$ gradually closer to the fixed image $I_f$.

## 2.3   Inter-patient Pretraining

As the provided training set is relatively small (140 image pairs), we pretrained our networks with inter-patient registration to avoid overfitting. Specifically, all images in the training set were shuffled and randomly paired, resulting in a total of 280×279 inter-patient image pairs. Then, the networks were pretrained with these inter-patient image pairs using a deep self-supervised loss $\mathcal{L}_{self}$ as follows:

$$\mathcal{L}_{self} = -\sum_{i=1}^{4} \frac{1}{2^{(4-i)}} NCC_w(I_m{}^i \circ \phi_i, I_f{}^i), \quad (1)$$

where the $NCC_w$ is the local normalized cross-correlation [16] with window size $w^3$. The $\mathcal{L}_{self}$ calculates the negative $NCC_w$ between the fixed and warped images at four registration steps, which measures the image similarity without ground truth labels.

In addition, we imposed L2 regularization on the $\phi_i$ to encourage its smoothness. Consequently, the total pretraining loss $\mathcal{L}_{pre}$ is defined as:

$$\mathcal{L}_{Pre} = \mathcal{L}_{self} + \sum_{i=1}^{4} \frac{1}{2^{(4-i)}} (\sum_{p \in \Omega} ||\nabla \phi_i(p)||^2). \quad (2)$$

## 2.4   Intra-patient Training with Dual Deep Supervision

After the inter-patient pretraining, our networks were further trained for intra-patient registration with the 140 intra-patient image pairs in the training set. In addition to the deep self-supervised loss $\mathcal{L}_{self}$ used during pretraining, we also introduced a deep weakly-supervised loss $\mathcal{L}_{weak}$ as follows:

$$\mathcal{L}_{weak} = \sum_{i=1}^{4} 2^{(4-i)} MSE(L_m{}^i \circ \phi_i, L_f{}^i), \quad (3)$$

where the $MSE$ is the mean square error between the coordinates of two sets of landmarks. The $\mathcal{L}_{weak}$ penalizes the distance between the fixed and warped landmarks at four registration steps, which leverages the information within the landmark labels to overcome the missing correspondences between pre-operative and follow-up scans.

In addition, as the landmark labels are sparse, the displacement fields should be smooth so that the landmark labels can impose more influence on the displacement fields. Therefore, in addition to the L2 regularization used during pretraining, we adopted an additional regularization loss that explicitly penalizes the negative Jacobian determinants [17], making the total regularization loss $\mathcal{L}_{reg}$ as:

$$\mathcal{L}_{reg} = \sum_{i=1}^{4} \frac{1}{2^{(4-i)}} [\lambda \cdot JD(\phi_i) + \sum_{p \in \Omega} ||\nabla\phi_i(p)||^2], \tag{4}$$

where the $JD$ is regularization loss penalizing the negative Jacobian determinants of $\phi_i$ (refer to [17]) and the $\lambda$ is a regularization parameter that adjusts the smoothness of displacement fields. Finally, the total training loss $\mathcal{L}_{train}$ is defined as:

$$\mathcal{L}_{Train} = \mathcal{L}_{self} + \mu\mathcal{L}_{weak} + \sigma\mathcal{L}_{reg}, \tag{5}$$

where the $\mu$ and $\sigma$ are two parameters that adjust the magnitude of each loss term.

## 2.5 Pair-specific Fine-tuning

We fine-tuned the trained networks for each inferred image pair and then used the fine-tuned networks for inference. The loss used for pair-specific fine-tuning is:

$$\mathcal{L}_{Fine} = \mathcal{L}_{self} + \sigma\mathcal{L}_{reg}. \tag{6}$$

As ground truth labels are not required, pair-specific fine-tuning can be used for any unseen image pairs and this has been demonstrated to improve the registration accuracy of deep registration methods [14,18].

## 2.6 Implementation Details

Our method was implemented using Keras with a Tensorflow backend on a 12 GB Titan V GPU. We used an Adam optimizer with a batch size of 1. Our networks were (inter-patient) pretrained for 50,000 iterations with a learning rate of $10^{-4}$ and then were (intra-patient) trained for another 14,000 iterations (100 epochs) with a learning rate of $10^{-5}$. We trained four networks for four MRI sequences (T1, T1CE, T2, and FLAIR) and combined them by averaging their outputs. During inference, for each inferred image pair, we fine-tuned the trained networks for 20 iterations with a learning rate of $10^{-5}$. The $\sigma$ and $\mu$ were set as 1.0 and 0.01 to ensure that the $\mathcal{L}_{self}$, $\sigma\mathcal{L}_{reg}$, and $\mu\mathcal{L}_{weak}$ had close values, while the $\lambda$ and $w$ were optimized based on validation results. Our method achieved the best validation results when $\lambda=10^{-4}$ and $w=3.0$.

## 2.7 Evaluation Metrics

The registration accuracy was evaluated based on manually annotated landmarks in terms of Mean Absolute Error (MAE) and Robustness. MAE is calculated between the landmark coordinates in the pre-operative scan and in the warped follow-up scan, where a lower MAE generally indicates more accurate registration. Robustness is a successful-rate metric in the range of [0, 1], describing the percentage of landmarks improving their MAE after registration. In addition, the smoothness of the displacement field was evaluated by the number of negative Jacobian determinants (NJD). As the $\phi$ is smooth and invertible at the voxel $p$ where the Jacobian determinant is positive ($|J\phi(p)| > 0$) [19], a lower NJD indicates a smoother displacement field.

## 3 Results and Discussion

Our validation results are summarized in Table 1. We trained four networks with four MRI sequences (T1, T1CE, T2, and FLAIR) and report their validation results respectively. Among the four networks, the network using T1CE achieved the best validation results (MAE: 3.486; Robustness: 0.818), followed by the networks using T1 (MAE: 3.917; Robustness: 0.785), FLAIR (MAE: 4.127; Robustness: 0.787), and T2 (MAE: 4.156; Robustness: 0.748). We combined these networks by averaging their outputs, in which we first combined all four networks and then removed the networks with relatively lower validation results one by one, resulting in three network ensembles (T1/T1CE/T2/FLAIR, T1/T1CE/FLAIR, and T1/T1CE). We found that all three ensembles achieved better validation results and produced smoother displacement fields, which suggests that combining different MRI sequences improves brain tumor registration. Combining more MRI sequences consistently improved NJDs as averaging more displacement fields together naturally resulted in a smoother displacement field. However, combining more MRI sequences cannot guarantee better registration accuracy. Among different ensembles, the T1/T1CE ensemble achieved the best validation results (MAE: 3.392; Robustness: 0.827), while further combining T2 and FLAIR networks resulted in lower validation results. In addition, combining the T1/T1CE networks by 0.3/0.7-weighted averaging further improved the performance and achieved our best validation results (MAE: 3.387; Robustness: 0.831). Therefore, we submitted this ensemble to be evaluated on the testing set, which made us place 4[th] in the final testing phase (Score: 0.3544) [4]. We also attempted to train a single network with multi-channel image pairs (concatenating MRI sequences into multiple channels), but this approach resulted in worse validation results.

In addition, we performed an ablation study to explore the contributions of dual deep supervision and pair-specific fine-tuning. In the ablation study, all four MRI sequences were used, and the deep weakly-supervised loss $\mathcal{L}_{weak}$ and/or the pair-specific fine-tuning process were excluded. The MAE results of the ablation study are shown in Table 2. Since pair-specific fine-tuning was not performed on the training

**Table 1.** Validation results of our method using different MRI sequences.

| MRI sequence | MAE | Robustness | NJD |
|---|---|---|---|
| T1 | 3.917 | 0.785 | 94.70 |
| T1CE | 3.486 | 0.818 | 68.65 |
| T2 | 4.156 | 0.748 | 69.70 |
| FLAIR | 4.127 | 0.787 | 100.00 |
| T1/T1CE/T2/FLAIR | 3.445 | 0.826 | **0.30** |
| T1/T1CE/FLAIR | 3.432 | 0.826 | 0.65 |
| T1/T1CE | 3.392 | 0.827 | 2.65 |
| T1/T1CE (weighted by 0.3/0.7) | **3.387** | **0.831** | 3.60 |

**Bold**: the lowest MAE, the highest Robustness, and the lowest NJD are in bold.

**Table 2.** MAE results of the ablation study on the training and validation sets.

| $\mathcal{L}_{self}$ | $\mathcal{L}_{weak}$ | PSFT | Training set | Validation set |
|:---:|:---:|:---:|:---:|:---:|
| √ | × | × | 4.375 | 3.716 |
| √ | × | √ | / | 3.598 |
| √ | √ | × | **1.026** | 3.521 |
| √ | √ | √ | / | **3.445** |

**Bold**: the lowest MAE on each set is in bold. PSFT: pair-specific fine-tuning.

set, the corresponding results are missing in Table 2. We found that using both $\mathcal{L}_{self}$ and $\mathcal{L}_{weak}$ (i.e. dual deep supervision) resulted in lower MAE than merely using $\mathcal{L}_{self}$. This is because ground truth information (landmark labels) was leveraged through the $\mathcal{L}_{weak}$, which helped to find the existing correspondences between images and thus relieved the challenges introduced by missing correspondences. However, when $\mathcal{L}_{weak}$ was used, the training MAE (=1.026) became much lower than the validation MAE (=3.521), indicating heavy overfitting. This is attributed to the fact that the provided training set is small (140 pairs) and the landmark labels are sparse (6-20 landmarks per pair). We suggest that a larger, well-labeled (more landmarks) training set will contribute to better registration performance. Also, we found that pair-specific fine-tuning can consistently contribute to lower MAE. This is consistent with existing studies [14, 18] where pair-specific fine-tuning (also named test-specific refinement or instance-specific optimization) was demonstrated to improve the registration accuracy. Pair-specific fine-tuning can improve the network's adaptability to image shape/appearance variations because the network can have chances to adjust the learned weights for each unseen image pair during inference.

Our method has some limitations and we suggest better performance potentially could be obtained by addressing them. Firstly, we omitted affine registration as all MRI scans provided by the challenge organizers have been rigidly registered to the same anatomical template [4]. However, we found that the first-place team adopted additional affine registration and this dramatically improved their registration performance [20]. This suggests that there still exist large linear misalignments between the provided MRI scans, and therefore additional affine registration is required. Secondly, as manually annotated landmarks are expensive to acquire, the provided landmark labels are sparse and thus led to heavy overfitting. For this limitation, automatic landmark detection methods could be considered to produce additional landmark labels. Finally, we adopted NICE-Net to perform four steps of coarse-to-fine registration. However, this step number (four steps) was empirically chosen without full exploration. We suggest that performing more steps of coarse-to-fine registration might result in better registration performance.

## 4      Conclusion

We outline a deep learning-based deformable registration method for brain tumor sequence registration in the context of BraTS-Reg 2022. Our method adopts our recently proposed NICE-Net as the backbone to handle the large deformations between pre-operative and follow-up MRI scans. Dual deep supervision, including a deep self-supervised loss based on image similarity and a deep weakly-supervised loss based on manually annotated landmarks, is deeply embedded into the NICE-Net, so as to overcome the missing correspondences between the tumor in the pre-operative scan and the resection cavity in the follow-up scan. In addition, pair-specific fine-tuning is adopted during inference to improve the network's adaptability to testing variations. Our method achieved a competitive result on the BraTS-Reg validation set (MAE: 3.387; Robustness: 0.831) and placed 4th in the final testing phase (Score: 0.3544).